\documentclass[runningheads]{llncs}

\PassOptionsToPackage{table}{xcolor}
\usepackage{eccv}

\usepackage{eccvabbrv}

\usepackage{graphicx}
\usepackage{booktabs}
\usepackage{multirow}
\usepackage{amsmath}
\usepackage{amssymb}

\usepackage{array}
\usepackage{ragged2e}
\newcolumntype{L}[1]{>{\RaggedRight\arraybackslash}p{#1}}
\newcolumntype{R}[1]{%
  >{\raggedright\arraybackslash}p{#1}%
}

\usepackage{hyperref}

\usepackage[accsupp]{axessibility}

\usepackage{orcidlink}

\begin{document}

\title{Diffusion-Generated Image Watermarking: A Two-Axis Taxonomy and Three
Protocol-Bounded Case Studies%
\thanks{An English translation, with partial reorganization, of our article
originally published in Korean in the \emph{Journal of Broadcast Engineering},
Vol.~31, No.~4, pp.~687--699, July 2026~\protect\cite{lee2026jbe}. Reused with
permission of the Korean Institute of Broadcast and Media Engineers (KIBME);
reused figures and tables are identified in their captions.}}

\titlerunning{Taxonomy and Case Studies of Diffusion-Image Watermarking}

\author{Sung Ju Lee\inst{1}\orcidlink{0009-0003-0269-5483} \and
Nam Ik Cho\inst{1}\orcidlink{0000-0001-5297-4649}}

\authorrunning{S.~J.~Lee and N.~I.~Cho}

\institute{Department of ECE \& INMC, Seoul National University, Korea\\
\email{\{thomas11809,nicho\}@snu.ac.kr}}

\maketitle

\begin{abstract}
  Watermarking diffusion-generated images requires balancing provenance signals with image quality, robustness, and computational cost. This work organizes methods along two axes: insertion mechanism and primary signal-bearing representation, and formalizes a representative $z_T$-Fourier pipeline for verification and identification. We then use the taxonomy to structure three protocol-bounded case studies. The first examines associations among frequency integrity, detection, quality, and cropping behavior. The second revisits persistence under seed-linked and seed-independent editing and formulates a scoped Semantic Imprinting Hypothesis without claiming a localized carrier or causal mechanism. The third studies single-shot VAE-latent phase modulation, including its efficiency, regeneration robustness, and robustness--quality operating points. Finally, we separate four content-level attack families from model/pipeline adaptation, propose corresponding evaluation protocols and testable conjectures for parameter-tuning threats, and identify additional temporal extensions for video. These analyses do not establish a universal ranking; instead, they provide a framework for matched, protocol-aware comparisons of watermarking systems for diffusion-generated images.
\end{abstract}

\section{Introduction}
\label{sec:intro}

The rapid evolution of generative AI, most notably text-to-image Latent Diffusion Models (LDMs), has democratized high-quality visual content creation. While early defense research focused primarily on synthetic image detection, modern media protection faces a broader challenge: verifying whether content has been misappropriated or altered via prompt-driven inpainting and semantic editing. Traditional pixel-domain watermarks and imperceptible noise injection~\cite{barni1998dct,cox2008digital} fail in this paradigm. Because diffusion-based editors re-synthesize images using learned generative priors, they tend to suppress subtle watermark perturbations that are weakly coupled to perceptually salient content~\cite{zhao2024invisible, ni2025fragility}. This vulnerability requires embedding watermark signals directly into the generative process itself.

To address this, \emph{semantic watermarking} has emerged as a promising alternative, with the literature referring to these methods as either sampling-based or inversion-based, depending on the underlying design perspective. These methods embed a watermark signal directly into the diffusion model's initial Gaussian noise seed $z_T$. During generation, the noisy latent is progressively denoised into the VAE latent $z_0$ and finally decoded into the pixel space. Existing approaches manipulate frequency patterns via predefined rules~\cite{wen2023treering,ci2024ringid}, encode payloads while preserving Gaussian noise statistics~\cite{yang2024gaussian}, or combine inversion with iterative optimization to embed watermark signals~\cite{zhang2024zodiac}. By embedding the watermark into the initial noise and propagating it through the diffusion sampling process, semantic watermarking can remain detectable under generative re-synthesis~\cite{zhao2024invisible, lee2025hstr}. However, this paradigm introduces distinct trade-offs: improper frequency-spectrum manipulation can degrade generation quality, whereas optimization-heavy methods incur substantial computational overhead.

Rather than merely comparing the performance of individual methods, this work examines diffusion-generated image watermarking from the perspective of the conditions under which embedded signals remain recoverable throughout generation, attacks, and detection. We organize the design space along two orthogonal axes, \emph{insertion mechanism} and \emph{target representation space}, providing a structured framework for analyzing how design choices shape the trade-offs among robustness, visual quality, and computational efficiency. Within this framework, we clarify how latent-space strategies overcome the limitations of pixel-domain watermarking, identify their remaining challenges, and outline an expanded threat model for future generative media protection.

Our analysis focuses on diffusion-generated static images, covering representative state-of-the-art latent watermarking methods together with foundational pixel-domain baselines for reference. Non-diffusion generative models (e.g., GANs) are excluded from the scope, while video watermarking is discussed only where temporal dynamics extend the image-centric threat model. The remainder of this paper is organized as follows. \Cref{sec:taxonomy,sec:formal} introduce the proposed two-axis taxonomy and formal system model. \Cref{sec:freq,sec:imprint,sec:phase} present three case studies, under their respective evaluation protocols, on frequency integrity and cropping, watermark persistence under generative editing, and the efficiency and regeneration robustness of phase modulation in the VAE latent space, respectively. \Cref{sec:threat} separates attacks on existing watermarked images from adaptation of the model or pipeline, presents evaluation guidance and three testable conjectures for parameter tuning, and briefly discusses the extension to video. \Cref{sec:conclusion} concludes with key future research directions.

\begin{table}[t]
  \centering
  \caption{Taxonomy matrix of generative-image watermarking methods by
  insertion mechanism and primary signal-bearing representation
  (\cref{sec:taxonomy:principle}). Reproduced from~\cite{lee2026jbe}.}
  \label{tab:taxonomy}
  \small
  \setlength{\tabcolsep}{2pt}
  \begin{tabular}{@{}L{2.45cm}@{\hspace{0.3cm}}L{3.20cm}@{\hspace{0.3cm}}L{1.85cm}@{\hspace{0.3cm}}L{3.70cm}@{}}
    \toprule
    & \multicolumn{3}{c}{Mechanism} \\
    \cmidrule(){2-4}
    Representation & Training-free & Opt.-based & Learning-based \\
    \midrule
    Initial noise ($z_T$)
      & Tree-Ring~\cite{wen2023treering} (freq.); RingID~\cite{ci2024ringid}
        (freq.); G.Shading~\cite{yang2024gaussian} (spatial);
        HSTR/HSQR~\cite{lee2025hstr} (freq.); \newline
        WIND~\cite{arabi2025wind}
        (freq., seed)
      & ZoDiac~\cite{zhang2024zodiac} (freq.)
      & DiffuseTrace~\cite{lei2024diffusetrace} (spatial) \\
    \midrule
    Intermediate ($z_t$) & -- & ROBIN~\cite{huang2024robin} & -- \\
    \midrule
    VAE latent ($z_0$)
      & PhaseMark~\cite{lee2026phasemark} \newline(freq., phase)
      & FreqMark~\cite{guo2024freqmark} (freq.)
      & RoSteALS~\cite{bui2023rosteals} (spatial);\newline
      Latent Watermark~\cite{meng2025latentwm} (spatial) \\
    \midrule
    Pixel
      & Classical DCT/DWT;\newline
      DwtDctSvd~\cite{cox2008digital} (freq.)
      & --
      & Stable Signature~\cite{fernandez2023stable} \newline(spatial, VAE-FT);\newline
        RivaGAN~\cite{zhang2019rivagan} \newline(spatial, attention);\newline
        HiDDeN~\cite{zhu2018hidden} (spatial) \\
    \bottomrule
  \end{tabular}
\end{table}

\section{A Two-Axis Taxonomy for Generative-Image Watermarking}
\label{sec:taxonomy}

Existing literature often conflates embedding mechanisms with target representation spaces, blending labels such as \emph{learning-based}, \emph{initial-noise-based}, and \emph{pixel-domain} onto a single descriptive axis. Such categorization obscures whether two methods share an algorithmic pipeline or simply target the same underlying tensor. 
To resolve this ambiguity, we establish a two-dimensional taxonomy that disentangles the \textbf{insertion mechanism} (how the signal is embedded) from the \textbf{insertion representation space} (where the signal resides). This orthogonal coordinate system, summarized in \Cref{tab:taxonomy}, provides a systematic framework to map the design space and expose the technical trade-offs inherent in generative image watermarking.

\subsection{Insertion Mechanism}
\label{sec:taxonomy:mechanism}

The mechanism axis characterizes the operational procedure used to embed the watermark, categorized into three distinct paradigms:
\begin{itemize}
    \item \textbf{Training-free:} Inserts signals via fixed rules or closed-form mathematical transformations (e.g., Fourier frequency modulation) without requiring parameter updates or iterative optimization during embedding.
    \item \textbf{Optimization-based:} Leaves the generative model parameters frozen while solving an iterative, backpropagation-driven optimization over latent vectors or additive latent perturbations to satisfy a target watermark signal.
    \item \textbf{Learning-based:} Trains or fine-tunes model parameters (e.g., the VAE decoder or diffusion backbone weights) or introduces auxiliary neural network modules to implicitly embed and extract watermark payloads.
\end{itemize}
Note that these categories focus on the dominant embedding procedure; for instance, a training-free embedding method may still rely on inversion routines for watermark extraction.

\subsection{Insertion Representation Space}
\label{sec:taxonomy:space}

The representation axis identifies the primary computational stage targeted by the embedding design along the generative pipeline: initial Gaussian noise $z_T$, intermediate denoising states $z_t$, clean VAE latents $z_0$, or output pixels. Where applicable, methods are further subdivided into \emph{frequency} or \emph{spatial} domains. Rather than rigid partitions, these representations span a continuous computational spectrum along the generative process:
\begin{itemize}
    \item \textbf{Initial Noise ($z_T$):} Targets the starting random latent seed, allowing the embedded signal to propagate through the entire denoising trajectory.
    \item \textbf{Intermediate States ($z_t$):} Intervenes directly within the iterative denoising process (e.g., ROBIN~\cite{huang2024robin}, which applies adversarial optimization over intermediate time steps, $t \in [200, 300]$).
    \item \textbf{VAE Latent Space ($z_0$):} Modulates the denoised latent vector $z_0$ generated after diffusion, prior to or during VAE decoding.
    \item \textbf{Pixel Domain:} Embeds signals directly into the finalized RGB image.
\end{itemize}

\subsection{Assignment Principle: Primary Signal-Bearing Representation}
\label{sec:taxonomy:principle}

The key principle in categorizing a method is not to confuse the representations traversed along the pipeline with the space where the signal is actually embedded and extracted. For example, Stable Signature~\cite{fernandez2023stable} processes VAE latents $z_0$ during generation, but $z_0$ itself is not constrained to hold a watermark pattern. Instead, it fine-tunes the VAE decoder so that the signal is embedded and extracted in the pixel domain, placing it under \emph{Pixel / Spatial Domain / Learning-based}. 
Conversely, ZoDiac~\cite{zhang2024zodiac} operates post-hoc on generated pixels and $z_0$, yet its optimization explicitly enforces a Tree-Ring-like Fourier pattern in the inverted initial noise $z_T$. Thus, despite its post-hoc pipeline, it is assigned to \emph{Initial Noise ($z_T$) / Frequency Domain / Optimization-based}.

Furthermore, while spatial versus frequency domain boundaries are sharp for rule-based or explicit transform approaches, they become subtle in learning-based schemes. For instance, the attention-based encoder in RivaGAN~\cite{zhang2019rivagan} embeds watermark payloads into learned pixel distributions, which differs conceptually from classic spatial noise addition. Rather than reflecting a flaw in our taxonomy, this subtle distinction illustrates how the insertion mechanism and representation space inherently interact.

\subsection{Matrix Coverage and Open Gaps}
\label{sec:taxonomy:matrix}

As shown in \Cref{tab:taxonomy}, mapping existing methods onto this matrix highlights clear technical trends and open gaps.
First, direct optimization over the initial noise $z_T$ is uncommon due to the high computational cost of backpropagating gradients through the entire generation trajectory; instead, existing methods resort to training-free rules or intermediate-stage optimization. Second, targeting intermediate representations ($z_t$) via training-free rules or parameter learning remains underexplored. 
Finally, direct pixel-level optimization is vulnerable to generative re-synthesis~\cite{guo2024freqmark}, leading optimization-based methods to focus on latent representations.

\begin{figure}[t]
  \centering
  \includegraphics[width=\linewidth,pagebox=cropbox,trim=0 0 0 55,clip]
    {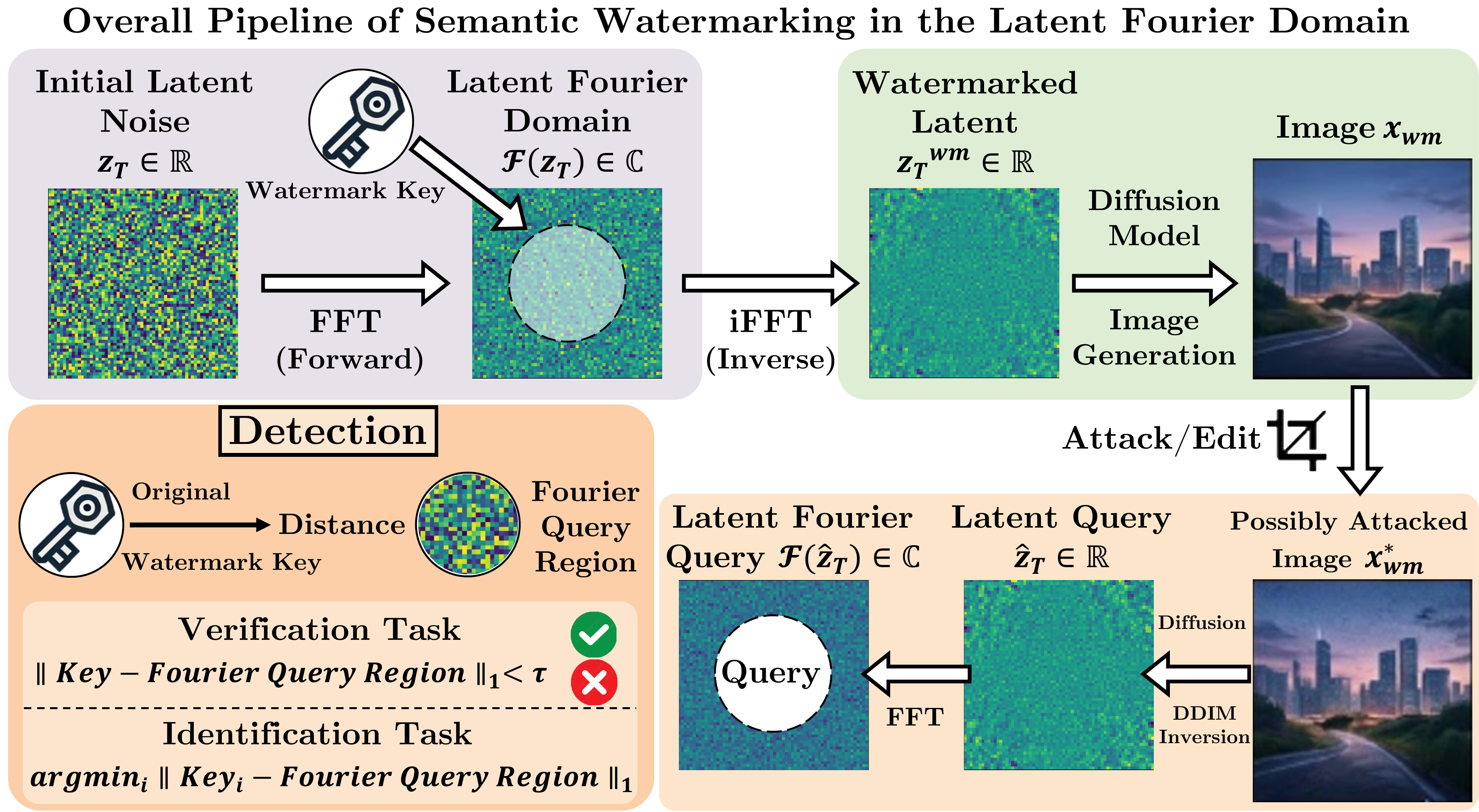}
  \caption{Representative generation-time $z_T$-Fourier watermarking pipeline
  used in the case studies: a keyed Fourier pattern is inserted into the
  initial noise, propagated through diffusion generation, recovered from an
  optionally transformed query via DDIM inversion, and compared with
  reference keys for verification or closed-set identification. Reproduced
  from~\cite{lee2026jbe}, which is itself adapted from~\cite{lee2025hstr}.
  The diagram does not represent post-hoc
  optimization, spatial-$z_T$, $z_0$, or pixel-domain methods, and the
  optional transformation is illustrative rather than a robustness claim.}
  \label{fig:pipeline}
\end{figure}

\section{Representative $z_T$-Fourier Watermarking: Setup and Detection Tasks}
\label{sec:formal}

Among the representation spaces in \cref{tab:taxonomy}, \cref{sec:formal,sec:freq,sec:imprint} adopt the $z_T$-Fourier family as a concrete case study to formalize semantic watermarking, examine frequency integrity, and analyze survival under generative editing. Evaluating this family requires definitions of the embedding and extraction pipelines as well as two detection tasks: verification (confirming watermark presence) and identification (attributing content to an enrolled user or model key). Below, we outline a representative pipeline; readout metrics and decision thresholds vary across specific methods. Note that both tasks are statistical decisions made under a fixed key and threshold, so neither constitutes proof of ownership on its own.

\subsection{A Representative $z_T$-Fourier Pattern Pipeline}
\label{sec:formal:embed}

The overall pipeline includes latent Fourier embedding, image generation, inversion-based extraction, and distance-based detection, as summarized in \cref{fig:pipeline}. Formally, a text-to-image diffusion model $G$ maps an initial Gaussian noise seed $z_T \sim \mathcal{N}(0, I)$ and a text prompt $p_s$ to a synthesized image $x = G(z_T, p_s)$. In $z_T$-Fourier semantic watermarking, a watermark pattern $w$ is embedded into a specified key region within the frequency domain $\mathcal{F}(z_T)$, yielding a watermarked seed $z_T^{\mathrm{wm}}$ and the corresponding watermarked image $x_{\mathrm{wm}} = G(z_T^{\mathrm{wm}}, p_s)$.

During detection, given a query image $x_{\mathrm{query}}$ that may have undergone downstream editing, an ODE-based DDIM inversion process reconstructs the inverted latent seed $\hat{z}_T$. Applying the Fourier transform to $\hat{z}_T$ enables the readout of the extracted key pattern $\hat{w}$ from the designated key region.

\subsection{Detection Tasks: Verification and Identification}

\paragraph{Verification.}
\label{sec:formal:verify}
Verification determines whether a specified watermark is present. In a standard distance-based formulation, the extracted pattern $\hat{w}$ is compared against a reference key $w$, typically $L_1$ distance in the Fourier domain:
\begin{equation}
  D_{\mathrm{verif.}}(x_{\mathrm{query}};w)
  = \mathbf{1}\!\left[d(\hat{w},w)<\tau\right],
  \label{eq:verification}
\end{equation}
where $\tau$ represents a predefined decision threshold calibrated on unwatermarked (negative) queries. Performance is evaluated using the True Positive Rate at a 1\% False Positive Rate (TPR@1\%FPR).

\paragraph{Closed-Set Key Identification.}
\label{sec:formal:identify}
Identification expands verification into a multi-class attribution task. It identifies which key from an enrolled pool $\{w_1, \ldots, w_N\}$ generated the query image under the assumption that a watermark exists. Under the closed-set $z_T$ pattern protocol, the predicted key index $\hat{i}$ is identified by finding the nearest reference key:
\begin{equation}
    \hat{i} = \arg\min_{i \in \{1, \ldots, N\}} d(\hat{w}, w_i).
    \label{eq:identify}
\end{equation}
Top-1 accuracy serves as the primary evaluation metric for the task. Because identification must distinguish the correct key among $N$ candidates rather than perform a simple presence test, it demands higher signal preservation under severe image distortions. In contrast to this pattern-matching formulation, conventional payload-based methods perform identification by applying source-specific thresholded decoding rules to recover the embedded message.

\section{Case Study I: Frequency Integrity, Detection, and Image Quality}
\label{sec:freq}

Early $z_T$ semantic watermarking schemes~\cite{wen2023treering, ci2024ringid} embed watermark signals by modifying specific frequency components of the initial noise vector. However, such spectral modifications often violate Hermitian symmetry in the Fourier domain. As a result, the imaginary components produced after the inverse Fourier transform must be discarded to enforce a real-valued spatial latent vector. This truncation disrupts the frequency integrity of the seed and shifts the latent distribution even further from the standard Gaussian prior. This compounded Out-of-Distribution (OOD) shift degrades both watermark detectability and the visual quality of generated images.

\subsection{Associations with Detection and Image Quality}
\label{sec:freq:integrity}

To resolve this limitation, HSTR and HSQR~\cite{lee2025hstr}, which we introduced in prior work, enforce Hermitian symmetry during frequency-domain embedding. This symmetry condition preserves seed integrity by ensuring lossless frequency reconstruction.

\begin{table}[t]
  \centering
  \caption{Detection over the harmonized attack subset and generative
  quality under the PhaseMark configuration. Except for HSTR, values
  follow~\cite{lee2026phasemark}; HSTR detection is
  recomputed from~\cite{lee2025hstr} without brightness and noise, and its
  $\Delta$FID-1k is our matched-condition measurement. ``Post'' denotes
  post-hoc applicability, and ``Payload/IDs'' reports payload bits and
  enrolled-pool size. Verification is TPR@1\%FPR. Identification follows
  each source-native rule: nearest-pattern accuracy for zero-bit methods or
  thresholded bit-decoding TPR@1\%FPR for payload methods. Lower
  $\Delta$FID-1k is better. Adapted from~\cite{lee2026jbe}.}
  \label{tab:detection}
  \setlength{\tabcolsep}{3pt}
  \begin{tabular}{@{}lccccc@{}}
    \toprule
    Method & Post & Payload / IDs & Verif. & Ident. & $\Delta$FID-1k \\
    \midrule
    DwtDctSvd~\cite{cox2008digital} & \checkmark & $32 / 10^6$ & 0.494 & 0.206 & $-0.389$ \\
    RivaGAN~\cite{zhang2019rivagan} & \checkmark & $32 / 10^6$ & 0.692 & 0.484 & $-0.785$ \\
    Stable Signature~\cite{fernandez2023stable} & $\times$ & $48 / 10^6$ & 0.819 & 0.489 & $-0.241$ \\
    Tree-Ring~\cite{wen2023treering} & $\times$ & zero / 2048 & 0.699 & 0.125 & $+1.184$ \\
    HSTR~\cite{lee2025hstr} & $\times$ & zero / 2048 & 0.993 & 0.946 & $-0.099$ \\
    RingID~\cite{ci2024ringid} & $\times$ & zero / 2048 & 0.999 & 0.977 & $+2.054$ \\
    HSQR~\cite{lee2025hstr} & $\times$ & zero / 8192 & 0.999 & 0.995 & $-0.721$ \\
    G.Shading~\cite{yang2024gaussian} & $\times$ & $256 / 10^6$ & 1.000 & 0.999 & $-0.317$ \\
    ZoDiac~\cite{zhang2024zodiac} & \checkmark & zero / 64 & 0.982 & 0.031 & $+0.091$ \\
    PhaseMark-PCQ~\cite{lee2026phasemark} & \checkmark & $128 / 10^6$ & 0.970 & 0.816 & $-0.517$ \\
    PhaseMark-APM~\cite{lee2026phasemark} & \checkmark & $128 / 10^6$ & 0.999 & 0.988 & $+0.042$ \\
    \bottomrule
  \end{tabular}
\end{table}
For the shared attack subset in \cref{tab:detection}~\cite{lee2025hstr,lee2026phasemark}, HSTR reaches verification of $0.993$ and identification of $0.946$. Tree-Ring reaches $0.699$ and $0.125$, respectively. HSTR combines Hermitian symmetry with insertion in the center region, so this comparison does not isolate the effect of symmetry.

RingID and HSQR both use structured binary patterns and reach a verification TPR at 1\% FPR of about $0.999$. Their image quality differs: HSQR records $\Delta\text{FID-1k} = -0.721$, while RingID records $+2.054$. This contrast is consistent with a relation between frequency integrity and image quality when detection is already high.

\subsection{Identification under Spatial Cropping}
\label{sec:freq:crop}

Beyond frequency integrity, resilience to spatial distortions such as cropping is essential for practical deployment. Conventional schemes that apply the Fourier transform across the entire spatial latent vector often induce severe information loss under image cropping. To address this, HSTR and HSQR adopt a center-aware insertion strategy, which embeds watermark signals into the central region of the latent vector.

\begin{table}[t]
  \centering
  \caption{Closed-set identification accuracy under random-crop attacks at
  four crop scales; smaller scales indicate more aggressive
  cropping~\cite{lee2025hstr}. Adapted from~\cite{lee2026jbe}.}
  \label{tab:crop}
  \setlength{\tabcolsep}{8pt}
  \begin{tabular}{@{}lcccc@{}}
    \toprule
    Method & 0.6 & 0.5 & 0.4 & 0.3 \\
    \midrule
    RingID~\cite{ci2024ringid} & 0.971 & 0.919 & 0.774 & 0.559 \\
    HSTR~\cite{lee2025hstr} & 1.000 & 0.999 & 0.992 & 0.903 \\
    HSQR~\cite{lee2025hstr} & 1.000 & 1.000 & 0.999 & 0.999 \\
    \bottomrule
  \end{tabular}
\end{table}

As reported in \cref{tab:crop}, HSTR and HSQR maintain high identification accuracy even under aggressive cropping ratios, whereas global-embedding methods like RingID suffer substantial performance drops. These results imply that localized insertion with frequency integrity helps maintain robustness against spatial cropping attacks.

\section{Case Study II: Persistence under Generative Editing}
\label{sec:imprint}

As noted in \Cref{sec:intro}, semantic watermarks show promising robustness against generative regeneration attacks~\cite{lee2025hstr}. Inversion-based editing such as Prompt-to-Prompt (P2P)~\cite{hertz2023p2p} remains anchored to the trajectory linking the source image to the initial seed $z_T$. Similarly, diffusion regeneration~\cite{zhao2024invisible} can be approximated as traversing this trajectory, since it adds forward noise to the source image to reach an intermediate latent state $z_t$ before denoising. Because these operations rely on the original latent trajectory, they are categorized as \emph{seed-linked}. Consequently, semantic watermarks may remain detectable when their structure is preserved in these recovered latent states.
However, recent findings show that watermarks also survive \emph{seed-independent} editing protocols such as InfEdit~\cite{xu2024infedit}, which synthesize images from newly sampled initial noise while retaining structural guidance. This survival challenges purely seed-linked explanations and leaves the precise watermark carrier unresolved.

\subsection{Observed Persistence across Editing Protocols}
\label{sec:imprint:results}

\begin{table}[t]
  \centering
  \caption{Verification TPR@1\%FPR after Prompt-to-Prompt (P2P) and InfEdit
  editing~\cite{lee2026imprinting}. Watermarked source images are generated
  with SD2.1, whereas InfEdit uses SD1.5, yielding a cross-model editing
  condition. This editing-specific protocol differs from the harmonized
  general-attack protocol in \cref{tab:detection}, so absolute values are not
  directly comparable across the two tables. Category labels are
  study-specific descriptors rather than classes in
  \cref{tab:taxonomy}. Reproduced from~\cite{lee2026jbe}.}
  \label{tab:editing}
  \setlength{\tabcolsep}{10pt}
  \begin{tabular}{@{}llcc@{}}
    \toprule
    Category & Method & P2P & InfEdit \\
    \midrule
    Spectral Modification & Tree-Ring~\cite{wen2023treering}     & 0.25 & 0.32 \\
    Optimization-based & ZoDiac~\cite{zhang2024zodiac}        & 0.75 & 0.86 \\
    Frequency-aligned & HSQR~\cite{lee2025hstr}              & 0.99 & 1.00 \\
    Cryptographic & G.Shading~\cite{yang2024gaussian}    & 1.00 & 0.98 \\
    \bottomrule
  \end{tabular}
\end{table}

\Cref{tab:editing} shows that HSQR and Gaussian Shading retain TPRs of at least $0.98$ under both editors, whereas Tree-Ring reaches $0.25$ under P2P and $0.32$ under InfEdit, as reported in our prior work~\cite{lee2026imprinting}.

Because P2P and InfEdit differ in algorithmic design beyond seed dependence, their numerical gap does not isolate the effect of replacing the original seed. 
In addition, detailed evaluations reveal that Tree-Ring already has low detectability on unedited source images, and this performance remains largely unchanged after editing. Since Tree-Ring directly modifies initial latent frequencies without preserving spectral integrity, its embedded pattern forms an out-of-distribution (OOD) signal that fails to align with the natural generative manifold. This structural limitation hinders reliable signal recovery even prior to editing, indicating that its low post-edit performance reflects poor initial detectability rather than edit-induced signal degradation.

If watermark recovery depended solely on reproducing the exact original seed, survival after editing initialized from newly sampled random noise would be impossible. However, because InfEdit remains source-conditioned, the high resilience of HSQR and Gaussian Shading suggests that well-aligned watermarks are internalized into the image feature space, including semantic structure, texture, or local statistics, and propagate reliably through the editing pipeline even without seed conservation.

\subsection{The Semantic Imprinting Hypothesis}
\label{sec:imprint:hypothesis}

The results support the hypothesis introduced in that work~\cite{lee2026imprinting}.

\begin{quote}
\textbf{Semantic Imprinting Hypothesis (SIH).} \emph{Under a fixed, source-conditioned editing protocol, a watermark remains detectable even after seed-independent editing if it is reliably recoverable from the unedited source image and its signal is carried within features preserved by the editor.}
\end{quote}

\begin{figure}[t]
  \centering
  \includegraphics[width=\linewidth]{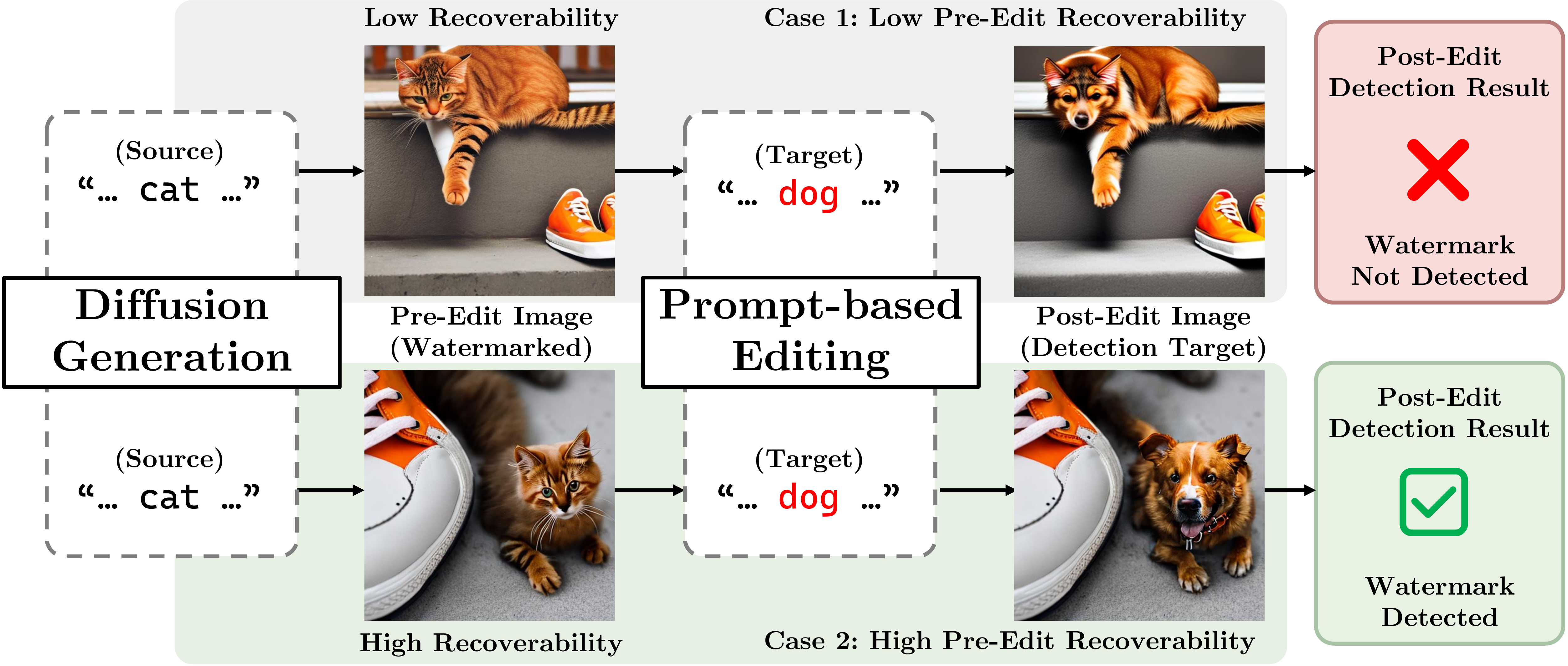}
  \caption{Schematic outcomes used to illustrate SIH. The upper example has
  low pre-edit recoverability and is not detected after editing, while the
  lower example has high pre-edit recoverability and is detected after
  editing. The figure does not identify the watermark carrier or show that
  recoverability before editing determines the detection result. Adapted
  from~\cite{lee2026jbe}; the underlying study is~\cite{lee2026imprinting}.}
  \label{fig:imprinting}
\end{figure}

\Cref{fig:imprinting} contrasts the failure and survival cases described by SIH according to pre-edit watermark recoverability.
The current evaluation does not directly test the proposed feature carrier because it covers one benchmark, one comparison between Stable Diffusion versions, and four methods. A stronger test should compare the same images before and after editing and report pre-edit TPR, post-edit TPR, and conditional retention $\Pr[D(\mathrm{Edit}(x))=1\mid D(x)=1]$ at a fixed threshold. It should control for editor type, edit strength, and fidelity. The test should also define a feature measure or a direct intervention. If recoverability before editing does not predict conditional retention under these controls, the hypothesis would be weakened.

\section{Case Study III: Single-Shot VAE-Latent Phase Modulation}
\label{sec:phase}

Watermarks in $z_T$ can be robust, but detection requires DDIM inversion and several seconds per image. The post-hoc $z_T$ method ZoDiac~\cite{zhang2024zodiac} shares this detection cost and also requires several minutes of optimization for embedding. PhaseMark~\cite{lee2026phasemark}, which we proposed in prior work, avoids DDIM inversion and embedding optimization. It modifies the phase of an intermediate frequency band in $z_0$ in a single step.

\subsection{Latency and Regeneration Robustness}
\label{sec:phase:latency}

\begin{table}[t]
  \centering
  \caption{Embedding procedure or per-image cost, detection latency, and mean
  verification TPR@1\%FPR across VAE-B, VAE-C, and diffusion
  regeneration~\cite{lee2026phasemark}. ``Sampling'' denotes generation-integrated
  insertion; ``opt.'' and ``inv.'' denote per-image optimization and DDIM
  inversion. Reproduced from~\cite{lee2026jbe}.}
  \label{tab:latency}
  \setlength{\tabcolsep}{4pt}
  \begin{tabular}{@{}lcccc@{}}
    \toprule
    Method & Space & Embedding & Detection & Regen.\ avg. \\
    \midrule
    ZoDiac~\cite{zhang2024zodiac} & $z_T$ & ${\sim}7.3$ min (opt.) & $4.11$ s (inv.) & 0.958 \\
    G.Shading~\cite{yang2024gaussian} & $z_T$ & Sampling & $4.11$ s (inv.) & 0.999 \\
    HSQR~\cite{lee2025hstr} & $z_T$ & Sampling & $4.11$ s (inv.) & 0.997 \\
    PhaseMark-APM~\cite{lee2026phasemark} & $z_0$ & $0.14$ s (one-shot) & $0.05$ s & 0.998 \\
    \bottomrule
  \end{tabular}
\end{table}

\Cref{tab:latency} shows that PhaseMark requires less than a second for both embedding and detection while maintaining an average regeneration verification rate ($0.998$) close to that of $z_T$-based semantic watermarks. This high survival rate demonstrates that regeneration robustness does not require initial noise seeds and is achievable through VAE latent frequency modulation.

\subsection{Modulation Strategies and Robustness--Quality Trade-offs}
\label{sec:phase:tradeoff}

\begin{table}[t]
  \centering
  \caption{PhaseMark variants: mean verification and identification
  TPR@1\%FPR over clean data and nine attacks, with image-quality
  metrics~\cite{lee2026phasemark}. Higher PSNR and lower LPIPS are better.
  Reproduced from~\cite{lee2026jbe}.}
  \label{tab:phase}
  \setlength{\tabcolsep}{6pt}
  \begin{tabular}{@{}llcccc@{}}
    \toprule
    Modulation & Strategy Type & Verif. & Ident. & PSNR\,$\uparrow$ & LPIPS\,$\downarrow$ \\
     & & (T@1\%F) & (T@1\%F) & & \\
    \midrule
    APM (Robust) & Hard / Absolute & 0.999 & 0.988 & 31.715 & 0.129 \\
    IPS & Hard / Relative & 0.998 & 0.980 & 32.437 & 0.090 \\
    SPS & Soft / Relative & 0.996 & 0.960 & 32.952 & 0.068 \\
    PCQ (Quality) & Soft / Absolute & 0.970 & 0.816 & 34.156 & 0.037 \\
    \bottomrule
  \end{tabular}
\end{table}

\Cref{tab:phase} shows a robustness--quality trade-off across the evaluated phase modulation strategies. Compared to Absolute Phase Modulation (APM), Phase Constellation Quantization (PCQ) yields lower detection scores but higher PSNR and lower LPIPS values. APM applies a stronger modulation to maximize signal detection, whereas PCQ preserves higher visual fidelity. This trade-off provides flexibility to choose a strategy based on application requirements.

\section{Threat Surfaces: Content Attacks and Model/Pipeline Adaptation}
\label{sec:threat}

The analyses in \cref{sec:freq,sec:imprint,sec:phase} evaluate attacks on watermarked images under a fixed embedding and detection pipeline. They do not cover changes to a released model or pipeline. We group threats into five families: signal processing, cropping, generative editing, regeneration, and model/pipeline adaptation. The first four transform existing images, while the last changes the system that produces or embeds future watermarks. We study the last family through parameter tuning. The families may overlap.

\subsection{Attack Families and Applicability Boundary}
\label{sec:threat:families}

\begin{table}[t]
  \centering
  \caption{Operational threat families and evidence scope. Content-level
  attacks transform an existing watermarked image under a fixed embedding
  rule, key, and detector; model/pipeline adaptation changes future watermark
  behavior and is supported here only by method-specific reports. Adapted
  from~\cite{lee2026jbe}.}
  \label{tab:attacks}

  \setlength{\tabcolsep}{4pt}
  \renewcommand{\arraystretch}{1.08}

  \begin{tabular}{@{}R{2.65cm}R{4.85cm}R{3.85cm}@{}}
    \toprule
    Attack family & Modified object & Evidence in this paper \\
    \midrule

    \rowcolor{black!7}
    \multicolumn{3}{@{}l@{}}{%
      \strut\textbf{Content-level attacks}\enspace
      {\scriptsize\color{black!65}
      Fixed embedding rule, key, and detector}%
    } \\
    \addlinespace[2pt]

    Signal processing
      & Samples of an existing image
      & General attacks~\cite{lee2025hstr,lee2026phasemark} \\
    \\

    Cropping
      & Spatial support of an existing image
      & Crop analysis~\cite{lee2025hstr} \\

    Generative editing
      & Source-conditioned re-synthesis
      & P2P; cross-model InfEdit~\cite{lee2026imprinting} \\

    Regeneration
      & Re-synthesis by VAE or LDM
      & Regeneration study\newline
        \cite{zhao2024invisible,ni2025fragility,lee2026phasemark} \\

    \addlinespace[2pt]
    \midrule
    \addlinespace[1pt]

    \addlinespace[2pt]
    \rowcolor{black!7}
    \multicolumn{3}{@{}l@{}}{%
      \strut\textbf{Model/pipeline-level adaptation}\enspace
      {\scriptsize\color{black!65}
      Future watermark behavior}%
    } \\
    \addlinespace[2pt]

    Adaptation
      & Generator, adapter, conditioning module, or recovery pipeline
      & Method-specific reports\newline
        \cite{feng2024aqualora,wang2025sleepermark} \\

    \bottomrule
  \end{tabular}
\end{table}

In the first four families in \cref{tab:attacks}, the embedding rule, key, and detector remain fixed, even if editing or regeneration uses another generative model. Parameter tuning changes the released generator, conditioning module, or integrated embedding component. Its future outputs are tested with the original key and detector. This threat applies when watermarking is integrated into the released system and remains active after tuning. If embedding is an optional external step, an operator can omit it without changing the model, so adaptation robustness does not cover that bypass.

Attack labels alone do not determine whether pixels, $z_0$, or $z_T$ will fail. Our representation taxonomy (\cref{tab:taxonomy}) classifies where a watermark is embedded, whereas its actual vulnerability depends on the protected object, pipeline integration, attacker access, and success criteria.
W-Bench evaluates watermark robustness under image regeneration and global and local generative editing~\cite{lu2025wbench}. Evidence for model/pipeline adaptation remains limited to individual methods. Accordingly, we focus on parameter tuning.

\subsection{Evaluation Protocols and Conjectures for Parameter Tuning}
\label{sec:threat:adaptation}

Full fine-tuning, LoRA~\cite{hu2022lora}, DreamBooth~\cite{ruiz2023dreambooth}, and Textual Inversion~\cite{gal2023textualinversion} cannot be placed on a universal severity scale based only on the number of parameters. Full fine-tuning may update broad portions of a generator; LoRA introduces low-rank adapter updates; DreamBooth is a personalization setting where trainable components vary across implementations; and Textual Inversion learns a conditioning embedding while leaving the base generator fixed. Any fair comparison must state the exact trainable tensors, objective, data, optimization budget, active embedding components, and detector model. The primary attack evaluation should keep the key, detector, and threshold calibrated on the original negative distribution fixed. Researchers may report recalibration on adapted outputs separately as a diagnostic for calibration shift.

AquaLoRA~\cite{feng2024aqualora} couples watermark behavior to U-Net adaptation and reports a trade-off between utility and removal in a white-box setting; however, this does not establish universal persistence under subsequent adaptation. SleeperMark~\cite{wang2025sleepermark} reports persistence under LoRA style adaptation, DreamBooth personalization, and ControlNet conditioning by separating watermark signals from learned semantic concepts. Because these findings apply only to individual methods rather than a matched ranking across \cref{tab:taxonomy}, they motivate testable conjectures instead of general conclusions.

\paragraph{Conjecture 1 (component overlap).}
Assume a watermark integrated into the system, where the operator cannot trivially detach or omit the embedding module. Here, persistence under adaptation relates more directly to whether the modified components overlap the watermark generation path than to the raw number of updated parameters.
\textbf{Testable by:} comparing updates that target the watermark generation path with those that do not, using matched prompts and seeds. Retained utility and adaptation objective performance should also be matched, and the status of the embedding module should be reported.

\paragraph{Conjecture 2 (inversion trajectory drift).}
An inversion dependent $z_T$ watermark can degrade indirectly when adaptation disrupts compatibility between the generation ($G$) and inversion ($I$) processes, even though its key is not stored in model weights. In contrast, direct latent level modulation ($z_0$) exhibits greater resilience against such trajectory distortion because it relies less on the inversion path.
\textbf{Testable by:} evaluating $(G_\theta,I_\theta)$, $(G_{\theta'},I_\theta)$, $(G_\theta,I_{\theta'})$, and $(G_{\theta'},I_{\theta'})$ with the same key, prompts, and seeds, and reporting latent recovery error alongside verification and identification. The fixed original inverter $I_\theta$ defines the primary attack result. Conditions using $I_{\theta'}$ provide diagnostics to separate changes in the forward model from inversion mismatches, rather than representing additional attacker privileges.

\paragraph{Conjecture 3 (adaptation interaction).}
Persistence depends on the interaction of adaptation type, objective, and strength. Consequently, training steps or the number of parameters alone do not produce a common severity ordering. 
\textbf{Testable by:} reporting detection, generation quality, and adaptation objective performance across attack strengths at matched retained utility. These evaluations characterize persistence and compatibility loss, not resistance to omission of the embedding step.

\subsection{Extension to Video}
\label{sec:threat:video}

Video watermarking introduces temporal attacks that do not occur in static images. VideoShield~\cite{hu2025videoshield} extends Gaussian Shading across time with template bits. It reports spatiotemporal tamper localization and temporal position recovery under frame exchange, insertion, and deletion, but extraction still uses DDIM inversion. VideoMark~\cite{hu2025videomark} embeds PRC coded messages frame by frame and uses edit distance for temporal alignment. It addresses inversion error, temporal distortion, and changes in sequence length. Video evaluations should cover insertion, deletion, exchange, and misalignment in addition to the five families above. They should also test inversion drift after temporal adapter tuning.

\section{Conclusion and Open Challenges}
\label{sec:conclusion}

This paper organizes watermarking methods for images generated by diffusion models according to insertion mechanism and representation space. It also formalizes verification and closed set identification for a representative Fourier watermark in $z_T$. The three case studies use different evaluation protocols, so their results do not provide a direct ranking. The results suggest a relation among frequency integrity, detection, and image quality, but the comparison does not separate Hermitian symmetry from other design changes. HSTR and HSQR also show higher identification accuracy than RingID in the crop tests. The editing results motivate the Semantic Imprinting Hypothesis, but the experiments do not identify which image features carry the watermark or test the proposed factors separately. PhaseMark requires less than one second for both embedding and detection and records an average verification rate of 0.998 in the regeneration tests. Its modulation methods show a trade-off between detection and image quality.

\Cref{sec:threat} separates attacks on an existing watermarked image from changes to the model or watermarking pipeline. Current evidence on adaptation comes from individual methods, so the paper does not rank these attacks by strength. The three conjectures concern component overlap, the relation between generation and inversion, and the effects of adaptation type, objective, and strength. Tests should state which components are trained, keep the key, detector, and threshold fixed, and report generation quality and performance on the adaptation task. Video also requires tests for frame insertion, deletion, exchange, and misalignment. Editing studies should test the same images before and after editing and report conditional retention at a fixed threshold. Adaptation studies should compare settings with similar generation quality and adaptation performance. These questions should also be tested with more diffusion models, watermark families, and video models. The verification and identification tasks in this paper do not by themselves establish legal ownership.

\section*{Acknowledgements}
This research was supported by Samsung Electronics Co., Ltd.

%
%
\bibliographystyle{splncs04}
\bibliography{refs}
\end{document}